\documentclass[11pt]{article}

\newif\ifcamera
\cameratrue

\ifcamera
  \usepackage[preprint]{neurips_2025}
\else
  \usepackage{neurips_2025}
\fi

\usepackage{times}
\usepackage{latexsym}

\usepackage[T1]{fontenc}
\usepackage[utf8]{inputenc}

\usepackage{microtype}
\usepackage{inconsolata}

\usepackage{amsfonts}
\usepackage{amsmath,amssymb}
\usepackage{nicefrac}

\usepackage{graphicx,verbatim}
\usepackage{booktabs}
\usepackage{multirow}
\usepackage{xcolor}
\usepackage{enumitem}
\usepackage[colorlinks=true,
            linkcolor=black,
            citecolor=blue,
            urlcolor=blue]{hyperref}

\def\keywordname{{\bfseries \emph Keywords}}%
\def\keywords#1{\par\addvspace\medskipamount{\rightskip=0pt plus1cm
\def\and{\ifhmode\unskip\nobreak\fi\ $\cdot$
}\noindent\keywordname\enspace\ignorespaces#1\par}}

\title{Evo-MedAgent: Beyond One-Shot Diagnosis with Agents That Remember, Reflect, and Improve}

\ifcamera
\author{%
Weixiang Shen$^{1,2,3,*}$ \quad Bailiang Jian$^{1,2,4,5,*}$ \quad Jun Li$^{1,5}$ \quad  \textbf{Che Liu}\textsuperscript{6} \quad Johannes Moll$^{1,2}$ \\ \textbf{Xiaobin Hu}\textsuperscript{4} \quad \textbf{Daniel Rueckert}$^{1,2,5,6}$ \quad \textbf{Hongwei Bran Li}$^{4}$ \quad \textbf{Jiazhen Pan}$^{1,2,5}$
  \\\\
  $^1$Technical University of Munich (TUM)\quad
  $^2$TUM University Hospital \quad
  $^3$LMU Munich \\
  $^4$National University of Singapore \quad
  $^5$Munich Center for Machine Learning (MCML)\\
  $^6$Imperial College London\\[0.2em]
$^{*}$Equal contribution \\ 
}

\else
\author{Anonymous Authors \\
Paper under double-blind review}
\fi

\begin{document}
\maketitle

\begin{abstract}
Tool-augmented large language model (LLM) agents can orchestrate specialist classifiers, segmentation models, and visual question-answering modules to interpret chest X-rays. However, these agents still solve each case in isolation: they fail to accumulate experience across cases, correct recurrent reasoning mistakes, or adapt their tool-use behavior without expensive reinforcement learning. While a radiologist naturally improves with every case, current agents remain static.
In this work, we propose Evo-MedAgent, a self-evolving memory module that equips a medical agent with the capacity for inter-case learning at test time. Our memory comprises three complementary stores: (1)~\emph{Retrospective Clinical Episodes} that retrieve problem-solving experiences from similar past cases, (2)~an \emph{Adaptive Procedural Heuristics} bank curating priority-tagged diagnostic rules that evolves via reflection, much like a physician refining their internal criteria, and (3)~a \emph{Tool Reliability Controller} that tracks per-tool trustworthiness. On ChestAgentBench, Evo-MedAgent raises multiple-choice question (MCQ) accuracy from 0.68 to 0.79 on GPT-5-mini, and from 0.76 to 0.87 on Gemini-3 Flash. With a strong base model, evolving memory improves performance more effectively than orchestrating external tools on qualitative diagnostic tasks. Because Evo-MedAgent requires no training, its per-case overhead is bounded by one additional retrieval pass and a single reflection call, making it deployable on top of any frozen model.
\end{abstract}

\begin{figure*}[t]
    \centering
    \includegraphics[width=\linewidth]{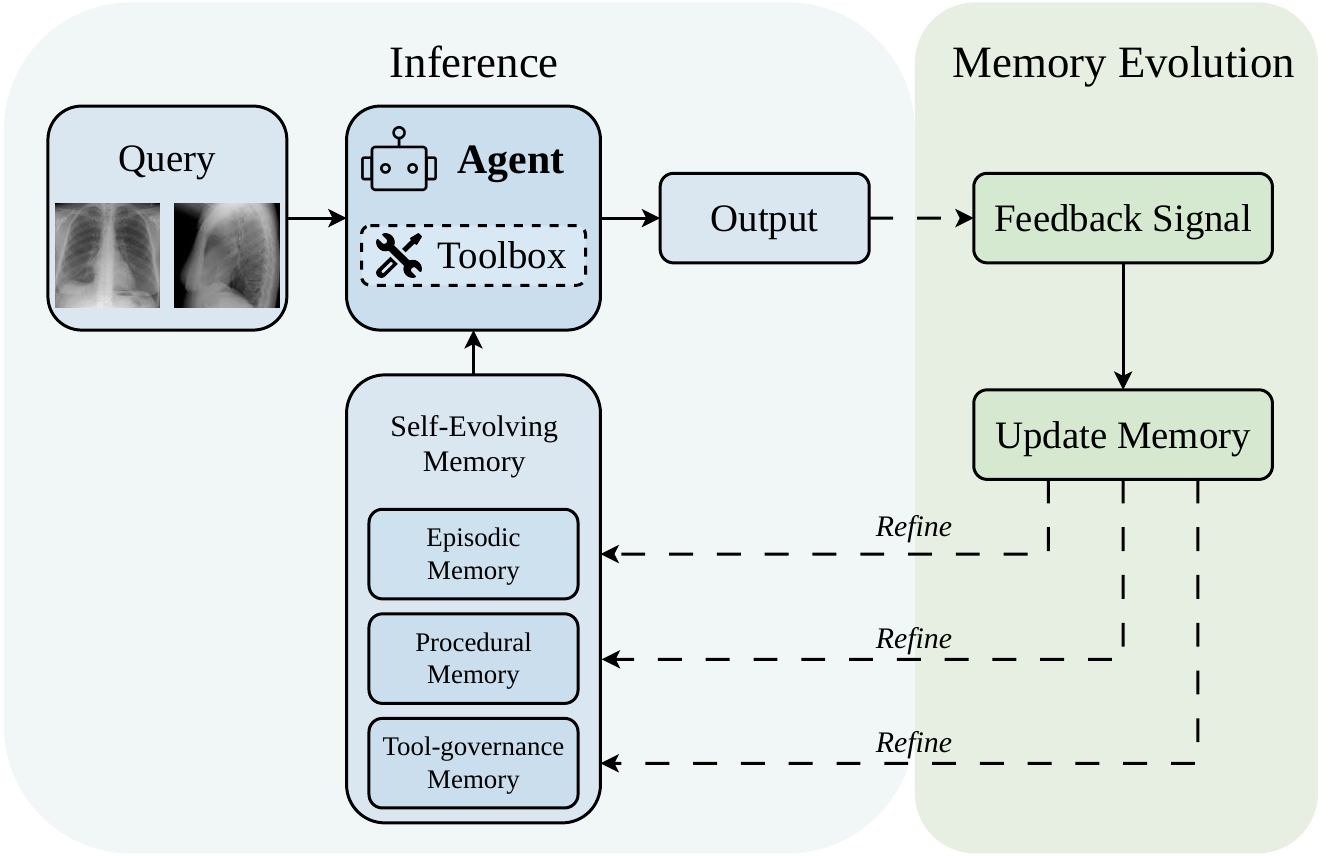}
    \caption{Overview of Evo-MedAgent. A tool-augmented LLM agent processes CXR questions sequentially while reading from and writing to a self-evolving memory via test-time learning. Before solving, the agent retrieves relevant episodic cases, procedural rules, and tool governance guidance. After receiving ground-truth feedback, it reflects and refines all three memory stores.}
    \label{fig:overview}
    \vspace{-10pt}
\end{figure*}

\section{Introduction}
\label{sec:intro}

Recent advances in large language models (LLMs) have enabled a new paradigm for medical image analysis: \emph{tool-augmented reasoning agents} that dynamically orchestrate specialist models to address complex clinical queries~\cite{fallahpour2025medrax,chen2024chexagent,zhu2025ask,zhao2026agentic}. Rather than relying on a single end-to-end model, these agents invoke classification, segmentation, visual question answering (VQA), phrase grounding, report generation, and dedicated vision-language models (e.g., LLaVA-Med~\cite{li2023llavamed}) as needed, combining their outputs through multi-step reasoning~\cite{chen2024huatuogpt,pan2025medvlm,gong2025med,liu2025beyond}. Systems such as MedRAX~\cite{fallahpour2025medrax} and CheXagent~\cite{chen2024chexagent} have demonstrated that this tool-augmented approach can improve performance on chest X-ray (CXR) interpretation tasks.

Despite their impressive capabilities, current medical agents lack inter-case memory~\cite{li2024mmedagent,kim2024mdagents,fallahpour2025medrax,pan2025beyond,thapa2025disentangling}. Each case is evaluated anew, with no mechanism to retain reusable lessons from prior successes and failures. An agent that misses a subtle pneumothorax on case~1 may repeat the same mistake on case~50 unless its parameters are retrained. Crucially, the agent's errors are as often procedural---wrong tool ordering, improper hyper-parameters, over-trust in a noisy classifier---as they are perceptual, yet no existing mechanism lets the agent systematically correct them. By contrast, radiologists refine their differential lists and search patterns with each case they read~\cite{ericsson2004deliberate}, learning which findings are commonly confused (e.g., left--right pneumothorax laterality), which tools to trust, and when to escalate uncertainty.

In this paper, we leverage a simple but effective test study in Chest X-rays (CXRs) and demonstrate that a medical reasoning agent can continually learn from its own clinical experience at test time. To achieve this, we introduce \textbf{Evo-MedAgent}, a self-evolving memory framework that can be built upon different medical VLM agents. This module performs in test-time and can be implemented in a plug-and-play manner without any post-training or weights fine-tuning.

Our central claim is that cross-case experience can be operationalized in a medically meaningful way: prior episodes provide precedent, distilled procedural memory carries reusable diagnostic policies, and tool-governance memory reshapes how the agent allocates trust across its toolset. This combination yields consistent, cumulative gains over no-memory systems. 

We make the following contributions:
\begin{enumerate}[leftmargin=*,itemsep=2pt]
    \item We present Evo-MedAgent, a \textbf{self-evolving test-time learning (TTL) memory} architecture and verify it in a chest X-ray-based medical diagnostic study. It evolves its agent memory from previous similar studies and elevates its performance from prior episodes, procedural heuristics, and tool-governance signals.
    \item We show that given a strong base model (e.g., GPT-5-mini), this evolving memory framework equipped without external tools delivers a stronger performance lift than orchestrating external specialist tools on qualitative diagnostic tasks.
    \item Evo-MedAgent is \textbf{training-free and low-overhead}, achieving continuous improvement without any LLM post-training: each case adds only one retrieval pass and one reflection call, enabling practical deployment on top of any frozen LLM.
\end{enumerate}

\section{Related Work}
\label{sec:related}

\textbf{Medical Reasoning Agents.}
Several recent systems frame CXR interpretation as multi-tool orchestration. MedRAX~\cite{fallahpour2025medrax} unifies seven specialist CXR tools (classification, segmentation, VQA, phrase grounding, report generation, CXR synthesis, and LLaVA-Med) under a ReAct-style~\cite{yao2022react} reasoning loop. AgentClinic~\cite{schmidgall2024agentclinic} simulates sequential clinical encounters in which a patient agent and doctor agent interact over multiple turns, but the agent's knowledge resets between encounters. MDAgents~\cite{kim2024mdagents} routes cases to specialist sub-agents by triage, yet each sub-agent operates statelessly. MedAgent-Pro~\cite{wang2025medagentpro} grounds each diagnostic step in retrieved medical evidence, though its evidence base is static and not updated from case outcomes. EHRAgent~\cite{shi2024ehragent} applies code-empowered reasoning to electronic health records but does not retain cross-patient memory. Hu et al.~\cite{hu2025landscape} survey this expanding landscape across modalities. All of these systems treat each case as an independent inference problem; none accumulates experience across cases.

\textbf{Memory and Knowledge Retrieval for LLM Agents.}
Standard retrieval-augmented generation (RAG) approaches, such as MMed-RAG~\cite{xia2024mmedrag}, provide static external knowledge at inference time. More broadly, agent memory has been explored through self-reflective feedback loops (Reflexion~\cite{shinn2023reflexion}), scalable long-term stores (Mem0~\cite{chhikara2025mem0}), and temporal knowledge graphs (Zep~\cite{rasmussen2025zep}), as well as multi-store architectures inspired by cognitive science (Generative Agents~\cite{park2023generative}). Recent test-time learning (TTL) works make this idea more explicit. Evo-Memory~\cite{wei2025evomem} proposes a three-store (episodic, semantic, procedural) architecture and evaluates it on general-purpose sequential tasks; we adopt a similar factorization but ground each store in radiological content---episodes record tool-interaction traces and case descriptors, procedures encode diagnostic heuristics, and governance tracks per-tool trust. Dynamic Cheatsheet~\cite{suzgun2025dyncheatsheet} maintains a single editable text block that an LLM rewrites after each task; we instead decompose this into priority-tagged procedural rules whose utility is tracked individually. ReMe~\cite{cao2025reme} refines procedural entries via a utility signal; we similarly employ utility-driven selection but additionally separate tool governance into its own store. Recent benchmarks such as CL-Bench~\cite{dou2026clbench} explicitly evaluate context-learning ability, while Git Context Controller~\cite{wu2025git} proposes version-controlled context management for LLM agents. To our knowledge, self-evolving test-time memory has not previously been evaluated in a tool-augmented medical imaging agent; the closest precedent is MMed-RAG~\cite{xia2024mmedrag}, which provides static retrieval rather than evolving stores.

\section{Methodology}
\label{sec:method}

We consider an ordered sequence of benchmark cases $(x_1, x_2, \ldots, x_N)$, where each case $x_i = (v_i, q_i)$ pairs chest X-ray image(s) $v_i$ with a question $q_i$ and has ground-truth answer $y_i$. Let $\mathcal{T}$ denote the available toolset. An agent produces a final answer $\hat{y}_i$ together with an interaction trace $\tau_i$ recording all tool calls, tool outputs, and intermediate reasoning. The agent operates sequentially, conditioned on a memory state $\mathcal{M}_{i-1}$ accumulated from preceding cases. After feedback $y_i$ is revealed, memory is updated:
\begin{equation}
    \mathcal{M}_i = f(\mathcal{M}_{i-1},\; x_i,\; \tau_i,\; \hat{y}_i,\; y_i).
    \label{eq:memory_update}
\end{equation}
The objective is sequential improvement without modifying the parameters of the underlying LLM or specialist tools. Our memory state is factorized as
\begin{equation}
    \mathcal{M}_i = (\mathcal{E}_i,\; \mathcal{S}_i,\; \mathcal{G}_i),
    \label{eq:memory_factorization}
\end{equation}
where $\mathcal{E}_i$ is episodic memory, $\mathcal{S}_i$ is procedural memory, and $\mathcal{G}_i$ is tool-governance memory. This factorization is motivated by prior multi-store memory formulations~\cite{park2023generative,wei2025evomem,cao2025reme}, but adapted to the requirements of medical image reasoning.

\subsection{Memory Architecture}
\label{sec:method:memory}

Our memory contains three stores with distinct state representations and update rules. Fig.~\ref{fig:overview} outlines the workflow.

\paragraph{Episodic memory ($\mathcal{E}$).}
The episodic store maintains compressed records of prior diagnostic episodes:
\begin{equation}
    e_j = (\phi_j,\; \tau_j,\; \hat{y}_j,\; y_j,\; \sigma_j,\; \gamma_j),
    \label{eq:episode}
\end{equation}
where $\phi_j$ is a compact case descriptor derived from $(v_j,q_j)$, $\tau_j$ is the tool-interaction trace, $\sigma_j$ is a one-sentence retrospective summary, and $\gamma_j$ is an actionable guideline distilled from reflection. At inference time, the agent retrieves the most relevant prior episodes:
\begin{equation}
    \mathcal{R}^{\text{epi}}_i = \operatorname{Top\text{-}K}_{e \in \mathcal{E}_{i-1}}\; s_{\text{epi}}(x_i, e),
    \label{eq:epi_retrieval}
\end{equation}
where $s_{\text{epi}}$ scores relevance between the current case and each stored episode. Retrieved episodes are injected as actionable precedents, allowing an observed failure or success to effectively constrain later reasoning.

\paragraph{Procedural memory ($\mathcal{S}$).}
Procedural memory stores reusable diagnostic policies distilled across episodes. Each policy $p_m = (u_m, \rho_m)$ pairs a short natural-language instruction $u_m$ with a priority level $\rho_m \in \{0, 1, 2\}$, where $\rho_m = 0$ denotes highest clinical urgency. The system tracks per-rule usage statistics (e.g., how often a rule was selected and whether the agent answered correctly when it was active) from which it derives a utility estimate $\mu(p)$ that favors rules with demonstrated benefit while providing an exploration bonus for under-tested ones. At inference time, up to $K_S$ policies are selected by a composite score combining relevance and utility.

\paragraph{Tool-governance memory ($\mathcal{G}$).}
In medical settings, learning when \emph{not} to trust a tool is critical. For each tool $t \in \mathcal{T}$, the governance store maintains a record $g_t = (\ell_t, n_t^{+}, n_t^{-}, n_t^{\text{mis}})$, where $\ell_t \in \{\textsc{Trusted}, \textsc{Caution}, \textsc{Avoid}\}$ is a trust label. We derive trust labels from fixed thresholds over accumulated interaction counts: a tool is labeled \textsc{Trusted} when its helpful rate exceeds $0.70$ with zero direct harm (on at least six interactions), \textsc{Avoid} when its effective-bad rate (harmful $+ 0.5\times$misuse, normalized) exceeds $0.60$ (on at least ten interactions), and \textsc{Caution} otherwise. Although tool governance is integral to Evo-MedAgent, we defer quantitative ablation of this component to future work and focus here on the episodic and procedural stores.

\subsection{Memory-Augmented Inference \& Update}
\label{sec:method:inference}

Before each case, the agent assembles a structured context prefix consisting of a fixed base protocol, governance guidance $\mathcal{R}^{\text{gov}}_i$, selected procedural policies $\mathcal{R}^{\text{proc}}_i$, and retrieved episodes $\mathcal{R}^{\text{epi}}_i$. After feedback $y_i$ is revealed, the agent performs multimodal reflection over the original images, question, response, tool traces, and ground truth to propose patch-style procedural edits and governance updates. Overall, each case follows a continuous cycle: reading from memory, reasoning about the case, receiving feedback, reflecting on errors, and writing the distilled insights back to the episodic, procedural, and governance memory stores. This enables true experience accumulation across the benchmark.

\section{Experiments}
\label{sec:experiments}

\subsection{Benchmark and Setup}
\label{sec:exp:benchmark}

\begin{table*}[t]
    \centering
    \caption{Ablation results on ChestAgentBench (tool-free). $\mathcal{S}$ = Procedural memory, $\mathcal{E}$ = Episodic memory. Full-memory GPT-5-mini MCQ is reported as mean (range) across three case-order permutations ($\mu{=}0.79$, $\sigma{=}0.025$). Dashes indicate incomplete runs. Best scores are marked in bold. }
    \label{tab:merged_results}
    \setlength{\tabcolsep}{11pt}
    \begin{tabular}{@{}lcc c cc c@{}}
        \toprule
        \multirow{2}{*}{Configuration} & \multicolumn{2}{c}{GPT-5-mini} & & \multicolumn{2}{c}{Gemini-3 Flash} & GPT-5.2 \\
        \cmidrule(lr){2-3} \cmidrule(lr){5-6} \cmidrule(lr){7-7}
         & MCQ & Open & & MCQ & Open & MCQ \\
        \midrule
        Baseline (Tool-Enabled) & 0.68 & -- & & 0.73 & -- & -- \\
        \midrule
        Baseline (Tool-Free) & 0.68 & 0.29 & & 0.76 & 0.21 & 0.69 \\
        + $\mathcal{S}$ (Procedural) & 0.71 & -- & & 0.81 & -- & -- \\
        + $\mathcal{E}$ (Episodic) & 0.78 & -- & & 0.84 & -- & -- \\
        Full Memory ($\mathcal{S}$+$\mathcal{E}$) & \textbf{0.79} (0.76--0.82) & 0.40 & & \textbf{0.87} & 0.38 & \textbf{0.81} \\
        \bottomrule
    \end{tabular}
\end{table*}

\begin{figure*}[h]
    \centering
    \begin{minipage}[t]{0.48\linewidth}
        \centering
        \includegraphics[width=\linewidth]{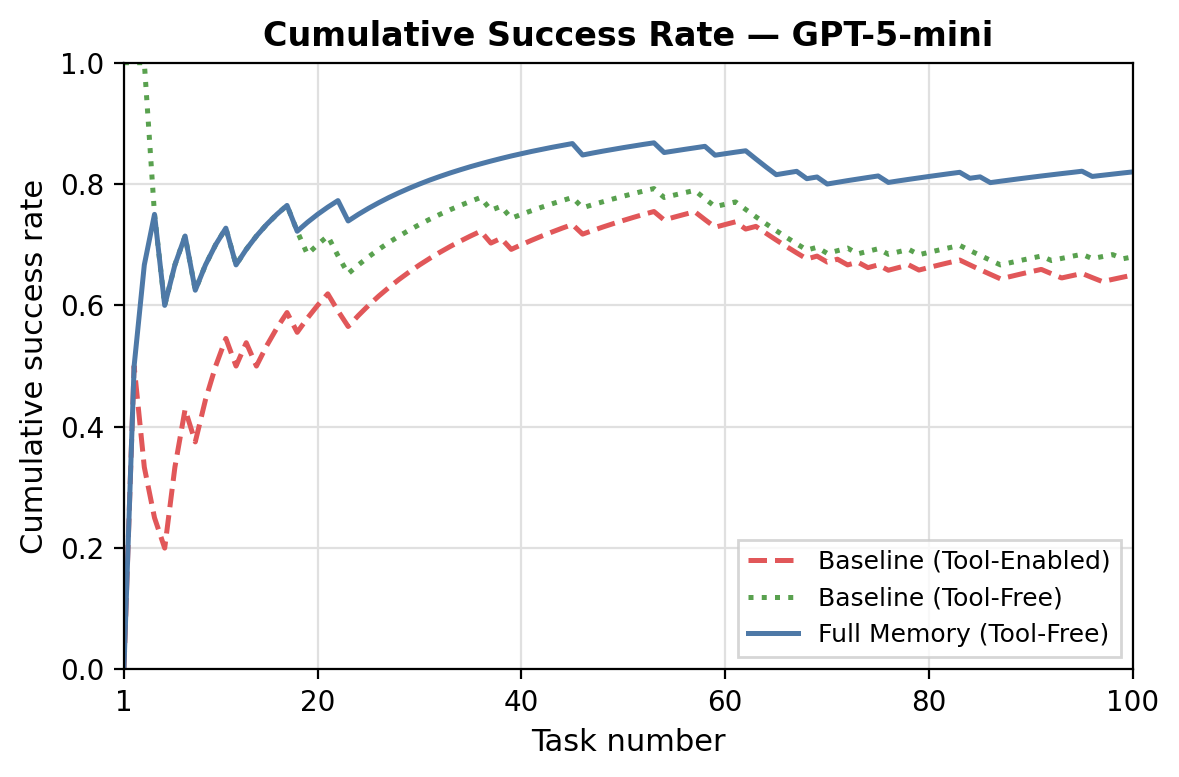}
    \end{minipage}
    \hfill
    \begin{minipage}[t]{0.48\linewidth}
        \centering
        \includegraphics[width=\linewidth]{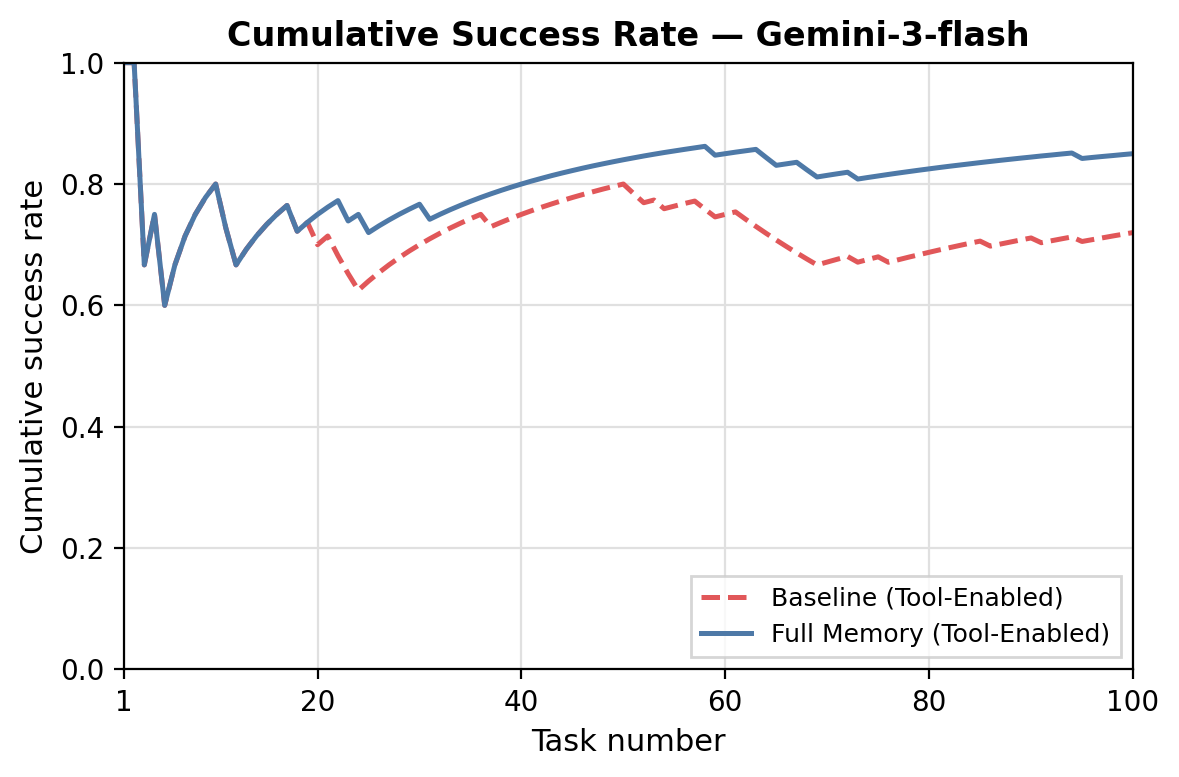}
    \end{minipage}
    \caption{Cumulative MCQ accuracy over sequential benchmark streams. \textbf{Left}: GPT-5-mini under tool-free baselines and full memory. \textbf{Right}: Gemini-3-Flash under tool-enabled baseline and full memory.}
    \label{fig:performance_curves}
\vspace{-10pt}
\end{figure*}

We evaluate the proposed framework on ChestAgentBench~\cite{fallahpour2025medrax}, a comprehensive benchmark containing 2{,}500 multiple-choice CXR questions across 675 expert-curated cases from the Eurorad database, spanning seven diagnostic categories: detection, classification, localization, comparison, relationship, characterization, and diagnosis. We study both tool-augmented and purely visual, tool-free regimes across multiple-choice and open-ended QA formats, using GPT-5-mini, Gemini-3 Flash, and GPT-5.2 as backbone VLMs.

\subsection{Experimental Configurations and Results}
\label{sec:exp:results}

Figure~\ref{fig:performance_curves} plots cumulative MCQ accuracy as cases are processed sequentially; Table~\ref{tab:merged_results} summarizes final accuracy for each ablation.

\textbf{Tool-Enabled vs.\ Tool-Free Capabilities.}
Evo-MedAgent's accuracy generally rises over the benchmark stream (illustrated for GPT-5-mini and Gemini-3 Flash in Figure~\ref{fig:performance_curves}), widening the gap to the no-memory baseline after approximately 20 cases. A key finding is that for qualitative diagnostic tasks, evolving memory provides a stronger performance liftup than using external specialist tools. When powered by a strong base VLM, the agent's internal reasoning, refined through inter-case experience, surpasses the tool-enabled baseline. For instance, on GPT-5-mini, the tool-free full-memory configuration achieves 0.79 mean accuracy, substantially outperforming the tool-enabled baseline (0.68). This suggests that for highly capable VLMs evaluating qualitative findings, the potential noise introduced by imperfect external localizers or segmenters is unnecessary; accumulated reasoning heuristics yield better, more consistent performance standalone. Corresponding improvements are also observed across open-ended evaluation formats (Table~\ref{tab:merged_results}).

\textbf{Division of Memory Labor.}
The ablation separating episodic from procedural memory reveals a division of labor: episodic retrieval, which surfaces similar past failures, dominates on recurring pathology patterns, while procedural heuristics generalize to novel pathology combinations unseen in recent episodes. This complementarity explains why the combined $\mathcal{S}$+$\mathcal{E}$ configuration consistently outperforms either component alone (Table~\ref{tab:merged_results}).

\textbf{System Robustness \& Order Sensitivity.}
Because test-time memory accumulates sequentially, sensitivity to case ordering is a methodological concern. We evaluated the full-memory configuration under multiple randomized permutations of the evaluation sequence. Across three random permutations, GPT-5-mini full-memory accuracy was 0.82, 0.76, and 0.79 ($\mu{=}0.79$, $\sigma{=}0.025$), indicating moderate robustness to case ordering.

\subsection{Qualitative Showcase}
\begin{figure*}[ht]
    \centering
    \includegraphics[width=0.85\linewidth]{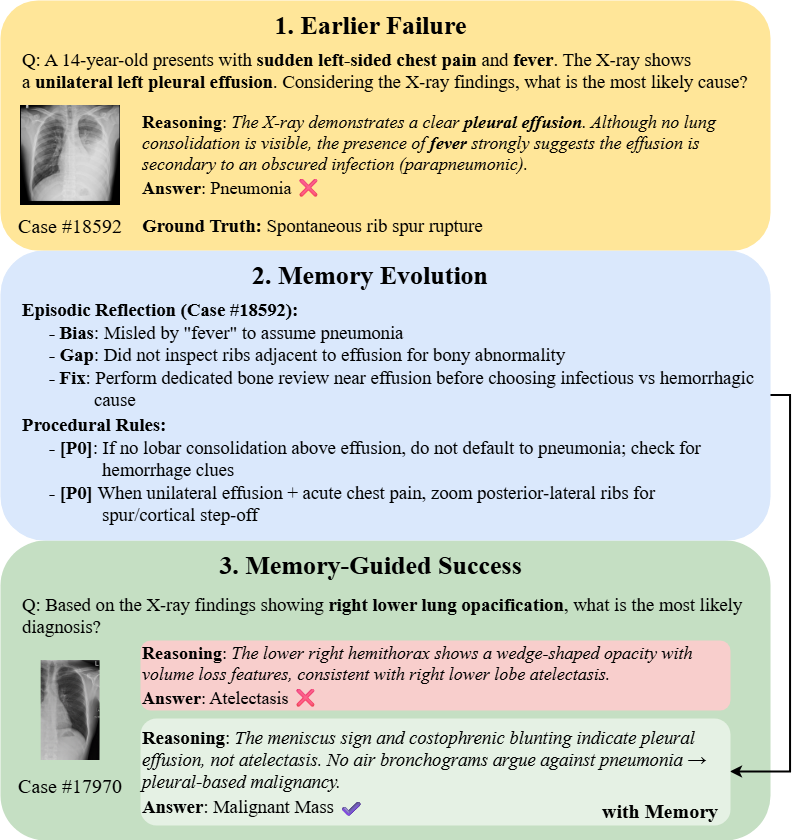}
    \caption{Evo-MedAgent showcase across two chest X-ray cases. \textbf{Earlier Failure}: the agent incorrectly attributes a left pleural effusion to pneumonia; ground truth is haemothorax from a rib spur. \textbf{Memory Evolution}: episodic reflection identifies the missed bone review and distils two new procedural rules. \textbf{Memory-guided Success}: on a later case, the without-memory agent misdiagnoses right lower lobe opacification as atelectasis, while the memory-augmented agent applies the retrieved rules (meniscus sign, absence of air bronchograms) to correctly identify a malignant pleural-based mass.}
    \label{fig:showcase}
\vspace{-10pt}
\end{figure*}

Figure~\ref{fig:showcase} illustrates the full Evo-MedAgent loop across two cases.
In Case~18592, a 14-year-old presents with sudden left-sided chest pain, fever, and a unilateral left pleural effusion.
Without memory the agent over-weights the fever, defaults to parapneumonic effusion, and answers \emph{Pneumonia}---missing the ground-truth diagnosis of haemothorax from a spontaneous rib spur.
Episodic reflection identifies the failure mode: the agent did not inspect the ribs adjacent to the effusion for bony abnormality.
Two procedural rules are distilled:
(i)~\emph{if no lobar consolidation is visible above an effusion, do not default to pneumonia; check for hemorrhage clues}, and
(ii)~\emph{when unilateral effusion co-occurs with acute chest pain, zoom posterior-lateral ribs for spur or cortical step-off}.

On a subsequent case (Case~17970), a chest X-ray shows right lower hemithorax opacification.
Without memory the agent interprets the opacity as a wedge-shaped volume-loss pattern and answers \emph{Atelectasis}.
With memory, the retrieved pleural-priority rules redirect attention to the meniscus sign and costophrenic blunting, while the absence of air bronchograms argues against pneumonia---leading to the correct answer of \emph{Malignant mass with pleural effusion} (primary pulmonary synovial sarcoma).

\section{Discussion and Conclusion}
\label{sec:conclusion}

\textbf{Limitations and Roadmap.}
This paper should be viewed as a \textbf{foundational first release} of Evo-MedAgent rather than a completed endpoint. The current study shows that self-evolving test-time memory can improve qualitative chest X-ray reasoning in a training-free manner, but several limitations remain. First, tasks requiring precise quantitative measurements, such as cardiothoracic ratio estimation or nodule sizing, still depend on deterministic specialist tools; in such settings, test-time memory should complement dedicated tool pipelines rather than replace them. Second, the present study focuses on an initial chest X-ray setting and assumes ground-truth feedback after each episode, whereas real clinical deployment often involves delayed, partial, or noisy supervision. Third, although tool-governance is a core component of our framework, its role is only introduced here and is not yet studied in sufficient depth. We therefore see this work as the beginning of a broader research program. Evo-MedAgent is under active development, and we are currently extending the project toward a \emph{v2 version} with more medical studies, more imaging modalities, and more datasets, together with a deeper investigation of tool-governance and how agents should calibrate trust, caution, and avoidance across heterogeneous tools. A more practical and straightforward extension would enable the proposed memory framework in MedOpenClaw~\cite{shen2026medopenclaw} for more efficient actions and tool-calling. 

\textbf{Conclusion.}
Evo-MedAgent provides initial evidence that a frozen medical agent can improve across cases through self-evolving memory, without model retraining. In this first release, we show that episodic and procedural memory can produce meaningful gains on ChestAgentBench and, for qualitative diagnostic tasks, can outperform a tool-enabled baseline when paired with a strong base VLM. More broadly, this work argues for a shift from one-shot medical agents toward systems that remember, reflect, and adapt over time. We view this paper not as the final form of the framework, but as a strong starting point for a continuing line of work toward broader clinical coverage, richer modalities, stronger tool-governance, and more realistic feedback settings.

\noindent\paragraph{Acknowledgements.} This work is partially funded by the European Research Council (ERC) project Deep4MI (884622).

\bibliographystyle{plain}
\bibliography{main}

@article{wei2025evomem,
  title={Evo-memory: Benchmarking llm agent test-time learning with self-evolving memory},
  author={Wei, Tianxin and Sachdeva, Noveen and Coleman, Benjamin and He, Zhankui and Bei, Yuanchen and Ning, Xuying and Ai, Mengting and Li, Yunzhe and He, Jingrui and Chi, Ed H and others},
  journal={arXiv preprint arXiv:2511.20857},
  year={2025}
}

@article{gong2025med,
  title={Med-CMR: A Fine-Grained Benchmark Integrating Visual Evidence and Clinical Logic for Medical Complex Multimodal Reasoning},
  author={Gong, Haozhen and Ji, Xiaozhong and Liu, Yuansen and Wu, Wenbin and Yan, Xiaoxiao and Liu, Jingjing and Wu, Kai and Pan, Jiazhen and Jian, Bailiang and Zhang, Jiangning and others},
  journal={arXiv preprint arXiv:2512.00818},
  year={2025}
}

@article{chen2024huatuogpt,
  title={Huatuogpt-o1, towards medical complex reasoning with llms},
  author={Chen, Junying and Cai, Zhenyang and Ji, Ke and Wang, Xidong and Liu, Wanlong and Wang, Rongsheng and Hou, Jianye and Wang, Benyou},
  journal={arXiv preprint arXiv:2412.18925},
  year={2024}
}

@inproceedings{pan2025medvlm,
  title={Medvlm-r1: Incentivizing medical reasoning capability of vision-language models (vlms) via reinforcement learning},
  author={Pan, Jiazhen and Liu, Che and Wu, Junde and Liu, Fenglin and Zhu, Jiayuan and Li, Hongwei Bran and Chen, Chen and Ouyang, Cheng and Rueckert, Daniel},
  booktitle={International Conference on Medical Image Computing and Computer-Assisted Intervention},
  pages={337--347},
  year={2025},
  organization={Springer}
}

@article{suzgun2025dyncheatsheet,
  title={Dynamic cheatsheet: Test-time learning with adaptive memory},
  author={Suzgun, Mirac and Yuksekgonul, Mert and Bianchi, Federico and Jurafsky, Dan and Zou, James},
  booktitle={Proceedings of the 19th Conference of the European Chapter of the Association for Computational Linguistics (Volume 1: Long Papers)},
  pages={7080--7106},
  year={2026}
}

@article{cao2025reme,
  title={Remember me, refine me: A dynamic procedural memory framework for experience-driven agent evolution},
  author={Cao, Zouying and Deng, Jiaji and Yu, Li and Zhou, Weikang and Liu, Zhaoyang and Ding, Bolin and Zhao, Hai},
  journal={arXiv preprint arXiv:2512.10696},
  year={2025}
}

@article{dou2026clbench,
  title={CL-bench: A Benchmark for Context Learning},
  author={Dou, Shihan and Zhang, Ming and Yin, Zhangyue and Huang, Chenhao and Shen, Yujiong and Wang, Junzhe and Chen, Jiayi and Ni, Yuchen and Ye, Junjie and Zhang, Cheng and others},
  journal={arXiv preprint arXiv:2602.03587},
  year={2026}
}

@article{wu2025git,
  title={Git context controller: Manage the context of llm-based agents like git},
  author={Wu, Junde and Hu, Minhao and Zhu, Jiayuan and Pan, Jiazhen and Liu, Yuyuan and Xu, Min and Jin, Yueming},
  journal={arXiv preprint arXiv:2508.00031},
  year={2025}
}

@inproceedings{fallahpour2025medrax,
  title={MedRAX: Medical Reasoning Agent for Chest X-ray},
  author={Fallahpour, Adibvafa and Ma, Jun and Munim, Alif and Lyu, Hongwei and Wang, Bo},
  booktitle={International Conference on Machine Learning},
  pages={15661--15676},
  year={2025},
  organization={PMLR}
}

@inproceedings{chen2024chexagent,
  title={Chexagent: Towards a foundation model for chest x-ray interpretation},
  author={Chen, Zhihong and Varma, Maya and Delbrouck, Jean-Benoit and Paschali, Magdalini and Blankemeier, Louis and Van Veen, Dave and Valanarasu, Jeya Maria Jose and Youssef, Alaa and Cohen, Joseph Paul and Reis, Eduardo Pontes and others},
  booktitle={AAAI 2024 Spring Symposium on Clinical Foundation Models},
  year={2024}
}

@inproceedings{li2023llavamed,
  title={Llava-med: Training a large language-and-vision assistant for biomedicine in one day},
  author={Li, Chunyuan and Wong, Cliff and Zhang, Sheng and Usuyama, Naoto and Liu, Haotian and Yang, Jianwei and Naumann, Tristan and Poon, Hoifung and Gao, Jianfeng},
  journal={Advances in Neural Information Processing Systems},
  volume={36},
  pages={28541--28564},
  year={2023}
}

@article{schmidgall2024agentclinic,
  title={Agentclinic: a multimodal agent benchmark to evaluate ai in simulated clinical environments},
  author={Schmidgall, Samuel and Ziaei, Rojin and Harris, Carl and Reis, Eduardo and Jopling, Jeffrey and Moor, Michael},
  journal={arXiv preprint arXiv:2405.07960},
  year={2024}
}

@inproceedings{shi2024ehragent,
  title={Ehragent: Code empowers large language models for few-shot complex tabular reasoning on electronic health records},
  author={Shi, Wenqi and Xu, Ran and Zhuang, Yuchen and Yu, Yue and Zhang, Jieyu and Wu, Hang and Zhu, Yuanda and Ho, Joyce C and Yang, Carl and Wang, May Dongmei},
  booktitle={Proceedings of the 2024 Conference on Empirical Methods in Natural Language Processing},
  pages={22315--22339},
  year={2024}
}

@article{kim2024mdagents,
  title={Mdagents: An adaptive collaboration of llms for medical decision-making},
  author={Kim, Yubin and Park, Chanwoo and Jeong, Hyewon and Chan, Yik S and Xu, Xuhai and McDuff, Daniel and Lee, Hyeonhoon and Ghassemi, Marzyeh and Breazeal, Cynthia and Park, Hae W},
  journal={Advances in Neural Information Processing Systems},
  volume={37},
  pages={79410--79452},
  year={2024}
}

@inproceedings{xia2024mmedrag,
  title={MMed-RAG: Versatile Multimodal RAG System for Medical Vision Language Models},
  author={Xia, Peng and Zhu, Kangyu and Li, Haoran and Wang, Tianze and Shi, Weijia and Wang, Sheng and Zhang, Linjun and Zou, James and Yao, Huaxiu},
  booktitle={The Thirteenth International Conference on Learning Representations}
}

@inproceedings{yao2022react,
  title={React: Synergizing reasoning and acting in language models},
  author={Yao, Shunyu and Zhao, Jeffrey and Yu, Dian and Du, Nan and Shafran, Izhak and Narasimhan, Karthik R and Cao, Yuan},
  booktitle={The eleventh international conference on learning representations}
}

@article{shinn2023reflexion,
  title={Reflexion: Language agents with verbal reinforcement learning},
  author={Shinn, Noah and Cassano, Federico and Gopinath, Ashwin and Narasimhan, Karthik and Yao, Shunyu},
  journal={Advances in neural information processing systems},
  volume={36},
  pages={8634--8652},
  year={2023}
}

@inproceedings{park2023generative,
  title={Generative agents: Interactive simulacra of human behavior},
  author={Park, Joon Sung and O'Brien, Joseph and Cai, Carrie Jun and Morris, Meredith Ringel and Liang, Percy and Bernstein, Michael S},
  booktitle={Proceedings of the 36th annual acm symposium on user interface software and technology},
  pages={1--22},
  year={2023}
}

@inproceedings{li2024mmedagent,
  title={Mmedagent: Learning to use medical tools with multi-modal agent},
  author={Li, Binxu and Yan, Tiankai and Pan, Yuanting and Luo, Jie and Ji, Ruiyang and Ding, Jiayuan and Xu, Zhe and Liu, Shilong and Dong, Haoyu and Lin, Zihao and others},
  booktitle={Findings of the Association for Computational Linguistics: EMNLP 2024},
  pages={8745--8760},
  year={2024}
}

@article{zhao2026agentic,
  title={An agentic system for rare disease diagnosis with traceable reasoning},
  author={Zhao, Weike and Wu, Chaoyi and Fan, Yanjie and Qiu, Pengcheng and Zhang, Xiaoman and Sun, Yuze and Zhou, Xiao and Zhang, Shuju and Peng, Yu and Wang, Yanfeng and others},
  journal={Nature},
  pages={1--10},
  year={2026},
  publisher={Nature Publishing Group UK London}
}

@inproceedings{zhu2025ask,
  title={Ask patients with patience: Enabling llms for human-centric medical dialogue with grounded reasoning},
  author={Zhu, Jiayuan and Pan, Jiazhen and Liu, Yuyuan and Liu, Fenglin and Wu, Junde},
  booktitle={Proceedings of the 2025 Conference on Empirical Methods in Natural Language Processing},
  pages={2846--2857},
  year={2025}
}

@article{shen2026medopenclaw,
  title={MedOpenClaw: Auditable Medical Imaging Agents Reasoning over Uncurated Full Studies},
  author={Shen, Weixiang and Hu, Yanzhu and Liu, Che and Wu, Junde and Zhu, Jiayuan and Shen, Chengzhi and Xu, Min and Jin, Yueming and Wiestler, Benedikt and Rueckert, Daniel and others},
  journal={arXiv preprint arXiv:2603.24649},
  year={2026}
}

@article{liu2025beyond,
  title={Beyond distillation: Pushing the limits of medical llm reasoning with minimalist rule-based rl},
  author={Liu, Che and Wang, Haozhe and Pan, Jiazhen and Wan, Zhongwei and Dai, Yong and Lin, Fangzhen and Bai, Wenjia and Rueckert, Daniel and Arcucci, Rossella},
  journal={arXiv preprint arXiv:2505.17952},
  year={2025}
}

@article{thapa2025disentangling,
  title={Disentangling reasoning and knowledge in medical large language models},
  author={Thapa, Rahul and Wu, Qingyang and Wu, Kevin and Zhang, Harrison and Zhang, Angela and Wu, Eric and Ye, Haotian and Bedi, Suhana and Aresh, Nevin and Boen, Joseph and others},
  journal={arXiv preprint arXiv:2505.11462},
  year={2025}
}

@article{pan2025beyond,
  title={Beyond benchmarks: Dynamic, automatic and systematic red-teaming agents for trustworthy medical language models},
  author={Pan, Jiazhen and Jian, Bailiang and Hager, Paul and Zhang, Yundi and Liu, Che and Jungmann, Friedrike and Li, Hongwei Bran and You, Chenyu and Wu, Junde and Zhu, Jiayuan and others},
  journal={arXiv preprint arXiv:2508.00923},
  year={2025}
}

@article{hu2025landscape,
  title={The Landscape of Medical Agents: A Survey},
  author={Hu, Xiaobin and Qian, Yunhang and Yu, Jiaquan and Liu, Jingjing and Tang, Peng and Ji, Xiaozhong and Xu, Chengming and Liu, Jiawei and Yan, Xiaoxiao and Yu, Xinlei and others},
  journal={Authorea Preprints},
  year={2025},
  publisher={Authorea}
}

@article{wang2025medagentpro,
  title={Medagent-pro: Towards evidence-based multi-modal medical diagnosis via reasoning agentic workflow},
  author={Wang, Ziyue and Wu, Junde and Cai, Linghan and Low, Chang Han and Yang, Xihong and Li, Qiaxuan and Jin, Yueming},
  journal={arXiv preprint arXiv:2503.18968},
  year={2025}
}

@article{chhikara2025mem0,
  title={Mem0: Building production-ready ai agents with scalable long-term memory},
  author={Chhikara, Prateek and Khant, Dev and Aryan, Saket and Singh, Taranjeet and Yadav, Deshraj},
  journal={arXiv preprint arXiv:2504.19413},
  year={2025}
}

@article{rasmussen2025zep,
  title={Zep: a temporal knowledge graph architecture for agent memory},
  author={Rasmussen, Preston and Paliychuk, Pavlo and Beauvais, Travis and Ryan, Jack and Chalef, Daniel},
  journal={arXiv preprint arXiv:2501.13956},
  year={2025}
}

@article{ericsson2004deliberate,
  title={Deliberate practice and the acquisition and maintenance of expert performance in medicine and related domains},
  author={Ericsson, K Anders},
  journal={Academic medicine},
  volume={79},
  number={Supplement\_2},
  pages={S70--S81},
  year={2004},
  publisher={Oxford University Press}
}

\end{document}